\documentclass[sigconf]{acmart}
\usepackage[ruled, linesnumbered, noline]{algorithm2e}
\usepackage{makecell}
\usepackage{array,multirow}
\usepackage{hyperref}
\usepackage{balance}

\AtBeginDocument{%
  \providecommand\BibTeX{{%
    \normalfont B\kern-0.5em{\scshape i\kern-0.25em b}\kern-0.8em\TeX}}}

\copyrightyear{2021}
\acmYear{2021}
\setcopyright{acmcopyright}\acmConference[CIKM '21]{Proceedings of the 30th ACM International Conference on Information and Knowledge Management}{November 1--5, 2021}{Virtual Event, QLD, Australia}
\acmBooktitle{Proceedings of the 30th ACM International Conference on Information and Knowledge Management (CIKM '21), November 1--5, 2021, Virtual Event, QLD, Australia}
\acmPrice{15.00}
\acmDOI{10.1145/3459637.3482215}
\acmISBN{978-1-4503-8446-9/21/11}

\begin{document}

\title{Variational Graph Normalized AutoEncoders}

\author{Seong Jin Ahn}
\email{sja1015@kaist.ac.kr}
\affiliation{%
  \institution{KAIST}
  \city{Daejeon}
  \country{Republic of Korea}
  \postcode{43017-6221}
}

\author{MyoungHo Kim}
\affiliation{%
  \institution{KAIST}
  \city{Daejeon}
  \country{Republic of Korea}}
\email{mhkim@dbserver.kaist.ac.kr}

\renewcommand{\shortauthors}{Short Paper Track}

\begin{abstract}
Link prediction is one of the key problems for graph-structured data.
With the advancement of graph neural networks, graph autoencoders (GAEs) and variational graph autoencoders (VGAEs) have been proposed to learn graph embeddings in an unsupervised way.
It has been shown that these methods are effective for link prediction tasks.
However, they do not work well in link predictions when a node whose degree is zero (i.g., isolated node) is involved.
We have found that GAEs/VGAEs make embeddings of isolated nodes close to zero regardless of their content features.
In this paper, we propose a novel Variational Graph Normalized AutoEncoder (VGNAE) that utilize $L_2$-normalization to derive better embeddings for isolated nodes.
We show that our VGNAEs outperform the existing state-of-the-art models for link prediction tasks.
The code is available at \hyperlink{https://github.com/SeongJinAhn/VGNAE}{https://github.com/SeongJinAhn/VGNAE}.
\end{abstract}



\maketitle

\section{Introduction}
Link prediction in the graph is to determine whether there is an unknown relationship between nodes in the incomplete original graph.
For example, it can be used to predict a link indicating if a specific substance in a biological network has a positive effect on the treatment of a specific disease, or if a specific literature or patent belongs to a certain author.

With the advancement of deep learning techniques, graph embedding models have been widely used in solving various problems in graphs.
Graph embedding models generate low-dimensional vectors for nodes in the graph, called node embeddings, and learn these vector representations according to their purpose. 
Among graph embedding technologies, Graph Convolutional Networks (GCNs) are known to outperform other graph embedding models \cite{kipf2016semi,defferrard2016convolutional,bruna2013spectral}.
GCNs utilize information of neighborhood to embed nodes by fusing features of the node itself and its neighbors. 
In link prediction tasks, a Graph Autoencoder (GAE) \cite{kipf2016variational} is used to learn node embeddings generated by the GCN-encoder via the reconstruction loss.
Since then, numerous GAE/VGAE variants have been proposed to achieve improved performance of link prediction \cite{pan2018adversarially, salha2020simple, mavromatis2020graph, di2020mutual}.

\begin{figure}[h]
    \centering
    \includegraphics[width=0.5\textwidth]{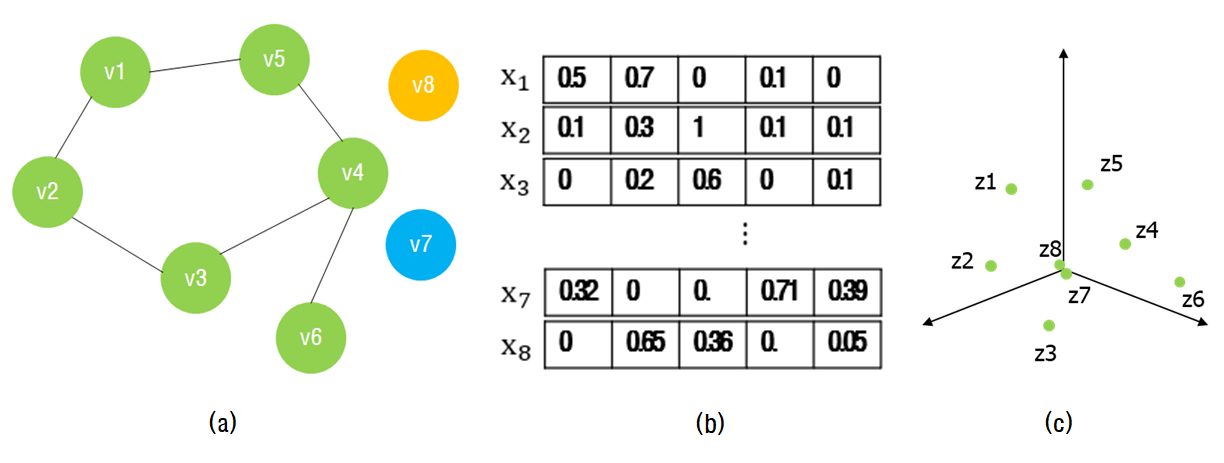}
    \caption{(a) A graph with isolated nodes ($v_7$ and $v_8$), (b) content features of nodes and (c) embedding space of latent vectors corresponding to node $v_i$ (i=1,2,...,8)}
    \label{fig:cluster}
\end{figure}

In many applications, there occur nodes in a graph that have no connection with other nodes.
In this paper, we call such nodes, i.e., nodes with no connection \textit{isolated nodes}.
For example, consider the following scenario.
There is a high school that maintains a social network G among members (e.g., students, professors, and staffs) where nodes are members and an edge between two members represents a "friendship" relation.
Suppose the school has a number of freshmen.
Here "Find out friends of students" can be considered a link prediction task where new nodes (freshmen) are involved.
Such new nodes do not have any connection initially, and hence are isolated nodes in G when link prediction tasks are performed.
Since there are no connectivity information of this case, feature contents of isolated nodes (e.g. the circles or hobby of students) play a major role in link prediction.

We have found that existing GAEs do not properly handle feature contents of isolated nodes.
GAEs learn to make a low similarity of embeddings between a pair of unconnected nodes.
Consider a graph consisting of eight nodes in Figure 1 (a). 
Figure 1 (c) shows vectors in an embedding space of the graph in Figure 1 (a) where $z_i$ (i=$1,2,...,8$) is a latent vector corresponding to node $v_i$ (i=$1,2,...,8$).
Relative positions of $z_i$ are determined based on the content information and the connectivity information in Figure 1 (a).
Now consider two isolated nodes $v_7$ and $v_8$.
Since $v_7$ and $v_8$ are not connected with other nodes, similarities between their embeddings and all other nodes should be low.
We have found that GAEs tend to make the Euclidean norm of embedding vectors of isolated nodes small in order to reduce similarities between embeddings of isolated nodes and all the other nodes.
As a result, the embeddings of isolated nodes go close to zero regardless of their content features.

In this paper, we propose a novel graph embedding technique, called Variational Graph Normalized AutoEncoder (VGNAE) for link prediction where the aforementioned problem of isolated nodes is properly handled.
We propose a Graph Normalized Convolutional Network (GNCN) that effectively use $L_2$-normalization to prevent embeddings of isolated nodes from going near zero.
Our VGNAE is a VGAE model where a GNCN is used to derive the mean and a GCN is used to derive the variance.
We show through extensive experiments that our proposed VGNAE effectively handles the problem of isolated nodes, and outperforms other existing state-of-the-art link prediction models.

\section{Preliminaries}
A graph $G$ can be represented as $G=(V,E,X)$ where $V$ is a set of vertices, $E \subset V \times V$ is a set of edges, and $X$ is a feature matrix of $V$.
\emph{N(v)} denotes a set of neighbors of $v \in V$, $n = |V|$ denotes the number of vertices, and $A \in R^{n \times n}$ is an adjacency matrix of $G$.
Let \emph{$\vec{z_v}$} be a vector that is an embedding of a node $v$, and \emph{$||\vec{x}||$} denotes an euclidean norm ($L_2$-norm) of vector $\vec{x}$.

\subsection{Graph Convolutional Networks}
Graph Convolutional Networks (GCNs) generalize the convolution operations to the graph domain.
The SpectralCNN \cite{bruna2013spectral} first proposes convolutional networks to the graph domains using the graph fourier transform.
The ChebyConv \cite{defferrard2016convolutional} parameterizes the graph convolution with chebyshev polynomials for efficient and localized filters.
The GCN \cite{kipf2016semi} simplifies ChebyConv by using a normalization trick.
Some unsupervised learning methods using GCNs have been proposed.
Kipf et al. \cite{kipf2016variational} propose two graph auto-encoders (GAEs and VGAEs) that reconstruct the adjacency matrix by node embeddings generated by GCNs.
LGAE \cite{salha2020simple} is a simple and interpretable linear models leveraging one-hop linear encoding.
ARGA and ARVGA \cite{pan2018adversarially} are two variants of adversarial approaches to learn robust embeddings.
GraphInfoClust \cite{mavromatis2020graph} captures richer information and nodal interaction by maximizing the mutual information between nodes of a same cluster.
sGraphite-VAE \cite{di2020mutual} extends the GNNs by exploring the aggregation using mutual information.

\subsection{$L_2$-normalization}
Certain properties about the norm of the embedding of the object ($||\vec{z_o}||$) have been addressed in several studies.
In neural translation models, an infrequent word is prone to have a embedding with a low $L_2$-norm ($||\vec{w}||$) \cite{kobayashi2020attention, arefyev2018much, schakel2015measuring, nguyen2017improving, nguyen2019transformers}.
In image recognition models, an embedding representing poor quality image has a low $L_2$-norm and vice versa \cite{liu2017sphereface,wang2018cosface}.
Also in image search, methods in \cite{wu2017multiscale, eghbali2019deep} normalizes the embedding to minimize the quantization error in high-resolution image search.
They and their subsequent studies use $L_2$-normalization \cite{ranjan2017l2,wang2017normface,zheng2018ring} to minimize errors caused by the imbalance between norms.
In addition, some works \cite{merrill2020parameter,nguyen2019transformers} show that the magnitude of the parameter continues (norm) to increase during gradient descent.
Zhang et al. \cite{zhang2020deep} turns out that the imbalance between norms causes an unstable direction update and uses $L_2$-normalization to resolve the problem.

As far as we know, PairNorm \cite{zhao2019pairnorm} and MSGNorm \cite{li2020deepergcn} are the only approach that use $L_2$-normalization in GCNs.
However, they are proposed to solve the over-smoothing problem, not for the problem caused by isolated nodes.

\section{Our Approach}
\subsection{Norm-zero tendency of isolated nodes}
For node $v \in \{v_1, v_2,..., v_n\}$ in a graph, there are certain relationships between the norm of nodes $||\vec{z_{v}}||$ from GAEs and degrees $d_v$.
Figure 2 (a) shows node embeddings from a GAE for the Cora and CiteSeer datasets in a 2-dimensional embedding space.
Figure 2 (b) shows the norms of node embeddings $||\vec{z_v}||$ from the GAE with respect to degrees of nodes for the Cora and CiteSeer datasets.
As shown in Figure 2 (a), embeddings of isolated nodes are around $\vec{0}$.
The norm of those vectors will be close to zero.
In Figure 2 (b), we can find out the norms of an embedding vectors of isolated nodes tend to be close to zero.
This also happens with the mean vector of VGAEs.
We call this phenomenon "norm-zero tendency of isolated nodes", which is an extreme case of the imbalance between norms.
This tendency makes embeddings of isolated nodes indistinguishable regardless of values of their content features.

\begin{figure}[h]
    \centering
    \includegraphics[width=0.45\textwidth]{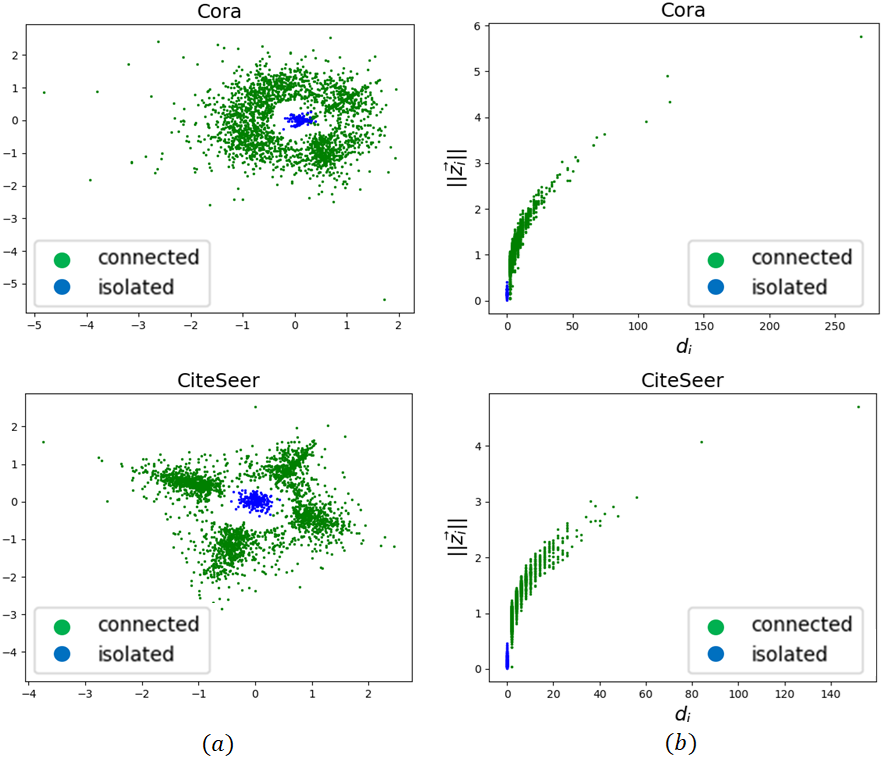}
    \caption{(a) A 2-dimensional embedding space of a GAE in the Cora and the CiteSeer dataset (b) a degree of nodes $d(v)$ and their norms of node embeddings $||\vec{z_v}||$ from a GAE in the Cora and the CiteSeer dataset.}
    \label{fig:cluster}
\end{figure}



\subsection{Graph Normalized Convolutional Network}

We propose a novel graph neural network called a Graph Normalized Convolutional Network (GNCN) that uses $L_2$-normalization before propagation.
Consider a feature matrix $X = [\vec{x_1}, \vec{x_2}, ... , \vec{x_n}]^T$ where $\vec{x_i} \in R^m$ is a content feature vector of node $v_i$ and $n$ is the number of nodes.  
A GNCN first generates the feature transformed vectors ($\vec{h} \in R^f$) with a learnable matrix W $\in R^{m \times f}$.
\begin{equation}
    \vec{h_i} = \vec{x_i}W
\end{equation}

Let $s \in R$ be a scaling constant that represents a norm of the hidden feature being propagated.
Our proposed GNCN generates the normalized feature transformed vectors ($\vec{n} \in R^f$) and propagates the normalized
vector to generates node embeddings ($\vec{z} \in R^f$).
\begin{equation}
    \vec{n_i} = s {\vec{h_i} \over ||\vec{h_i}||}
\end{equation}
\begin{equation}
    \vec{z_i} = {1 \over {d_i}+1} \vec{n_i} + \sum_{j\in N(i)}{1 \over \sqrt{d_i+1}\sqrt{d_j+1}} \vec{n_j} \quad {\rm where} \quad i \in \{1,2,...,n\}
\end{equation}

Now, for a feature matrix $X \in R^{n \times m}$ and an adjacency matrix $A$, $GNCN(X,A,s)$ is defined as follows:
\begin{equation}
    {\rm GNCN(X,A,s)} = s\tilde{D}^{-{1 \over 2}}\tilde{A}\tilde{D}^{-{1 \over 2}} g(XW)
\end{equation}

Here $g([\vec{h_1}, \vec{h_2}, ... , \vec{h_n}]^T) = [{\vec{h_1} \over ||\vec{h_1}||}, {\vec{h_2} \over ||\vec{h_2}||}, ..., {\vec{h_n} \over ||\vec{h_n}||}]^T$ , $Z = [\vec{z_1}, \vec{z_2}$ $, ... , \vec{z_n}]^T \in \mathbb{R}^{n \times f}$ is a node embedding matrix, $W \in \mathbb{R}^{m \times f}$ is a trainable matrix, $\tilde{A} = A+I_N$, $\tilde{D}$ is a degree matrix of $\tilde{A}$, and $s$ is a scaling constant. 
We will show that our proposed GNCN properly handles the norm-zero tendency of isolated nodes through experiments in Sec 4.4.1.

\subsection{Variational Graph Normalized AutoEncoder}

In this paper, we propose two variants of graph autoencoder called Graph Normalized AutoEncoder (GNAE) and Variational Graph Normalized AutoEncoder (VGNAE).
For each node $v \in \{1,2,...,n\}$, GNAE encodes the local structure information and node feature information of its neighborhood to derive latent variables $\vec{z_v} \in R^f$.
To generate $Z=[\vec{z_1},\vec{z_2},...,\vec{z_n}]^T\in R^{n \times f}$, GNAE uses a GNCN encoder that avoids the norm-zero tendency of isolated nodes.
GNAE uses a inner-product decoder to create the reconstructed adjacent matrix $\hat{A}$ from $Z$ as follows:
\begin{equation}
   \hat{A} = \sigma(Z^T Z), \quad {\rm with} \quad    Z = GNCN(X,A,s)
\end{equation}
where $\sigma$ is a sigmoid function.

We also propose a Variational Graph Normalized AutoEncoder (VGNAE).
Since mean vectors in VGAEs also have norm-zero tendency of isolated nodes, we derive mean vectors of VGNAE with a GNCN encoder.
Our VNGAE takes a simple inference model by using the mean field approximation to define the variational family as follows:
\begin{equation}
    q(Z|X,A) = \prod_{i=1}^n q(z_i | X,A) \quad {\rm with} \quad q(z_i | X,A) = N(z_i | \mu_i, diag({\sigma_i}^2))
\end{equation}
where $\mu = [\mu_1,\mu_2,...,\mu_n]^T = GNCN(X,A,c)$ is the matrix of mean vectors $\mu_i$ ; similarly $log \sigma = [log \sigma_1,...,log \sigma_i]^T = GCN(X,A)$.

Our generative model reconstructs graph structure $A$ by using a simple inner product decoder.
\begin{equation}
    p(A|Z) = \prod_{i=1}^n \prod_{j=1}^n p(A_{ij} | \vec{z_i},\vec{z_j}) \quad {\rm with} \quad  p(A_{ij}=1 | Z) = \sigma(\vec{z_i}^T \vec{z_j})
\end{equation}

For GNAE, we minimize the reconstruction error of $\hat{A}$.
For VGNAE, optimization is made by maximizing a tractable variational lower bound (ELBO) as follows:
\begin{equation}
    L_{ELBO} = E_{q(Z|X,A)}[log p(A|Z)] -  KL(q(Z|X,A) || p(Z))
\end{equation}
where $KL(q||p) = \Sigma_j Q_j log({Q_j \over P_j})$ is the Kullback-Leibler divergence between $q$ and $p$.
We use a Gaussian prior $p(Z)$ $=$ $\prod_{i=1}^n$ $N(z_i | 0,I)$.

\section{Experiments}
We conduct various experiments to show performance improvements of our proposed VGNAEs in link prediction when isolated nodes are involved.
First, we show the performance improvement of our GNCN by comparing the AUC scores of isolated nodes in various types of graph structured networks.
Then, we evaluate our GAE/VGNAE employing the GNCN.

\subsection{Datasets}
We use various types of attributed graphs.
First, we use three citation network datasets (Cora, Citeseer, and Pubmed).
Second, we use a coauthor network Coauthor CS \cite{shchur2018pitfalls}.
Third, we use a co-purchase graph Amazon Photo \cite{shchur2018pitfalls}.
Statistics about datasets are described in Table 1.
\begin{table}
  \caption{The statistics of the benchmark graph datasets}
  \label{sample-table}
  \centering
  \begin{tabular}{lllll}
    \toprule
    Dataset & Type  &  {\#}Nodes     & {\#}Edges  & {\#}Features \\
    \midrule
    \midrule
    Cora    & Citation  & 2,708  &  5,429  &  1,433     \\
    CiteSeer & Citation & 3,327   &  4,732  &  3,703    \\
    PubMed & Citation & 19,717  &  44,338 &  500 \\
    CS & Coauthor & 18,333 & 81,894 & 6,805 \\
    Photo & Co-purchase & 7,487 & 119,043 & 745 \\
    \bottomrule
  \end{tabular}
\end{table}

\subsection{Setup}
We implement all our models using Pytorch 1.4.0 \cite{paszke2019pytorch}.
We use the Adam optimizer \cite{kingma2014adam} with a learning rate of 0.005.
We train all models for a maximum of 300 epochs and early stopping with a window size of 50.
In all the experiments, 64 dimensions are used for node embeddings.
A scaling constant ($s$) in GNCN is set as 1.8.
For GCN \cite{kipf2016semi}, we use a two-layer GCN with the dimension of hidden embeddings is set to 128.
For GraphHeat \cite{xu2020graph} and APPNP \cite{klicpera2018predict}, the optimal hyper-parameters (e.g. scaling parameter $s$ and teleport rate $t$) are chosen through validation set.
For GraphHeat \cite{xu2020graph}, 0.4 is used for the coeffient $s$.
For APPNP \cite{klicpera2018predict}, 0.15 is used for the teleport rate and 10 is used for the number of propagations.
The link prediction models are evaluated by the area under the ROC curve (AUC) and average precision (AP) scores.

\subsection{Results}
\subsubsection{GNCN : Power of normalization for isolated nodes}
In this section, we present that performance of GAEs with existing GCN encoders degrades when isolated nodes are involved.
We also present that GAEs using our proposed GNCN encoders effectively encode isolated nodes.
We compare the performance of GAEs using various GCN-based encoders for isolated nodes and connected nodes on Cora, CiteSeer, CS and, Photo datasets.
The compared GCN-based encoders are GCN \cite{kipf2016semi}, GAT \cite{velivckovic2017graph}, SGCN \cite{wu2019simplifying}, SuperGAT \cite{kim2020find}, APPNP \cite{klicpera2018predict}, GraphHeat \cite{xu2020graph}, PairNorm \cite{zhao2019pairnorm}, and MSGNorm \cite{li2020deepergcn}.


For every dataset, we use a training set 60\%, a validation set 10\%, and a test set 30\% among all edges.
We add the same number of randomly sampled negative edges to the valid set and test set.
We measured the AUC score for isolated nodes and connected nodes in test sets.
The results are shown in Figures 3.
Figure 3 shows that the AUC scores of GAEs with other GCN encoders of isolated nodes is significantly lower than that of connected nodes (10 $\sim$ 20\%) for all types of graphs.
Experimental results on graphs of various types show that the degrade of performance in isolated nodes occurs in general graphs.
In addition, it can be seen that the accuracy at the isolated nodes for each GCN varies depending on the type of dataset.
We confirmed that existing GCNs are not suitable encoders of GAEs when isolated nodes are involved.
The methods using proposed GNCN ensured the highest performance of the isolated nodes for all graphs without compromising the performance of the connected nodes.

\begin{figure}[t]
    \centering
    \includegraphics[width=0.4\textwidth]{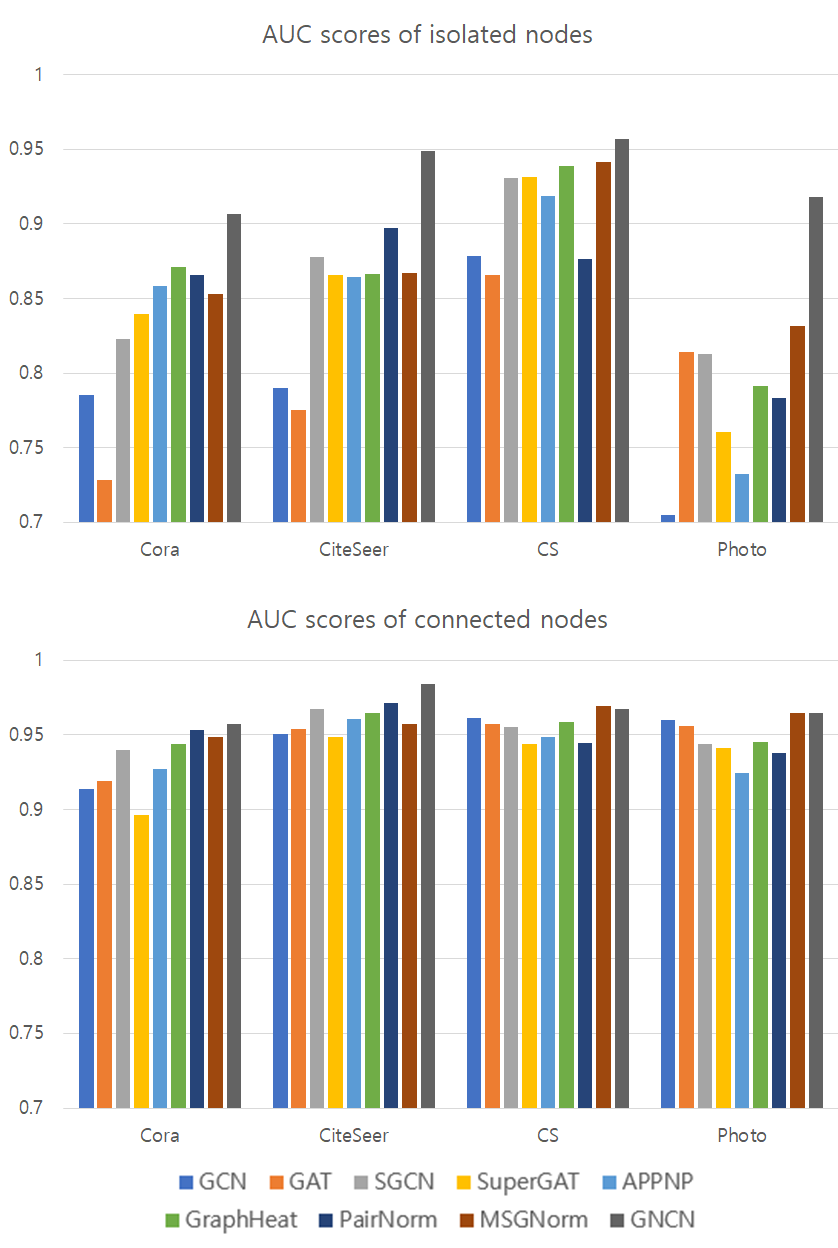}
    \caption{AUC scores of isolated nodes and connected nodes}
    \label{fig:cluster}
\end{figure}

\subsubsection{Performance Comparision of GNAE/VGNAE with state-of-the-art methods}
We conduct experiments on citation networks (Cora, CiteSeer, and PubMed) to compare the performance of GNAE/VGNAE with state-of-the-art link prediction models, i.e., LGAE \cite{salha2020simple}, ARGA \cite{pan2018adversarially}, ARGVA \cite{pan2018adversarially}, Graph InfoClust (GIC) \cite{mavromatis2020graph}, and sGraphite-VAE \cite{di2020mutual}.
For all datasets, 20\%, 40\%,  and 80\% of edges are used for training sets.
For the remaining edges, the ratio of 1 to 3 are used for valiatdation sets and test sets.
We add the same number of randomly sampled negative edges for each valid and test set.
For each dataset divided in this way, the AUC and AP scores are measured.
The results of link prediction in the dataset are shown in Table 2.

\renewcommand{\tabcolsep}{3pt}

\begin{table}[!h]
  \caption{Experiment results for citation datasets on link prediction task}
  \label{sample-table2}
  \centering
  {\small
  \begin{tabular}{ c c c c c c c c c c c }
    \hline
    Data    & Metric & Train & GAE & LGAE & ARGA  & GIC & sGraph & GNAE & VGNAE\\
    \midrule
    \multirow{6}{*}{\rotatebox[origin=c]{0}{Cora}}   
                            & \multirow{3}{*}{AUC}  
                             & 20\%  & 0.782  & 0.866 & 0.795 & 0.880 & 0.845 & 0.887 & \textbf{0.890}
                            \\ \cline{3-10} 
                             & & 40\%   & 0.856 & 0.908 & 0.844  & 0.914  & 0.840 &  0.926 & \textbf{0.929}
                             \\ \cline{3-10} 
                             & & 80\%  & 0.922  & 0.938  & 0.919   & 0.933   & 0.885  &  \textbf{0.956} & 0.954
                             \\ \cline{2-10}
                             & \multirow{3}{*}{AP}  
                              & 20\%  & 0.793  & 0.878 & 0.806  & 0.881  &  0.829  &  \textbf{0.901} & \textbf{0.901}
                             \\ \cline{3-10} 
                             & & 40\%  & 0.861 & 0.915 & 0.856  & 0.911  & 0.828  & \textbf{0.936} &  0.933              
                             \\ \cline{3-10} 
                             & & 80\%  & 0.930 & 0.945 & 0.927  & 0.929  & 0.867  &  0.957 & \textbf{0.958}
                             \\ \hline

\multirow{6}{*}{\rotatebox[origin=c]{0}{CiteSeer}}   
                            & \multirow{3}{*}{AUC}  
                              & 20\%  & 0.786  & 0.906 & 0.750  &  0.930  & 0.928  &  \textbf{0.946} & 0.941
                             \\ \cline{3-10} 
                             & & 40\%  & 0.836 & 0.925 & 0.832  & 0.936  & 0.936  & 0.956 & \textbf{0.961}
                             \\ \cline{3-10} 
                             & & 80\%  & 0.894  & 0.955 & 0.904 & 0.962  & 0.963  &  0.965 & \textbf{0.970}
                             \\ \cline{2-10}
                             & \multirow{3}{*}{AP}  
                             & 20\%  & 0.797  & 0.913 & 0.777  & 0.934 & 0.897  & \textbf{0.953} & 0.948
                             \\ \cline{3-10} 
                             & & 40\%  & 0.850  & 0.929 & 0.844  & 0.938  & 0.910  &  0.958 & \textbf{0.966}
                             \\ \cline{3-10} 
                             & & 80\%  & 0.903 & 0.959  & 0.915  & 0.966  & 0.943   & 0.970 & \textbf{0.971}
                             \\ \hline
                             
\multirow{6}{*}{\rotatebox[origin=c]{0}{PubMed}}   
                            & \multirow{3}{*}{AUC}  
                            & 20\%  & 0.937  & 0.946 & 0.936  & 0.950  &  0.837  & 0.950 & \textbf{0.951}
                            \\ \cline{3-10} 
                             & & 40\%  & 0.959 & 0.962 & 0.955  &  0.958 & 0.876 &  0.963  & \textbf{0.964}
                             \\ \cline{3-10} 
                             & & 80\%  & 0.967 & 0.974 & 0.973  & 0.960  &  0.896  & 0.975 & \textbf{0.976}
                             \\ \cline{2-10}
                             & \multirow{3}{*}{AP}  
                            & 20\%  & 0.940 & 0.947 & 0.941  & 0.947  &  0.859  & \textbf{0.950} & 0.949
                             \\ \cline{3-10} 
                             & & 40\%  & 0.961 & 0.961 & 0.959 & 0.956  & 0.879 &  0.961 & \textbf{0.963}
                             \\ \cline{3-10} 
                             & & 80\%  & 0.967 & 0.975 & \textbf{0.976}  & 0.965  & 0.902 & 0.975 & \textbf{0.976}
                             \\ \hline
    \bottomrule
  \end{tabular}
  }
\end{table}

As can be seen in Table 2, our GNAE/VGNAE show superior results compared to other methods in all divisions.
Also we can observe that the fewer observed edges (the smaller the ratio of the training rate), the better the performance of our proposed method compared to other SOTA models.
This is because as the ratio of unobserved edges increase, the number of isolated nodes also increases.

\section{Conclusions}
We have presented that in GAE and VGAE embeddings of isolated nodes tends to go to zero regardless of their content features on various graph-structured networks. 
This tendency can significantly degrade accuracy of link prediction when many isolated nodes are involved.
We have argued that the $L_2$-normalization is an effective technique to generate proper embeddings of isolated nodes in the link prediction task. 
We have shown through extensive experiments that our proposed VGNAE performs better than other existing methods.

\section*{Acknowledgements}
This work was supported by the Bio-Synergy Research Project of the MSIT (No. 2013M3A9C4078137), the National Research Foundation of Korea(NRF) grant (No. 2020R1A2C1004032), and the ITRC support program (IITP-2020-2020-0-01795) supervised by the IITP. 

\bibliographystyle{ACM-Reference-Format}
\balance 
\bibliography{VGNAE}


\begin{thebibliography}{32}


\ifx \showCODEN    \undefined \def \showCODEN     #1{\unskip}     \fi
\ifx \showDOI      \undefined \def \showDOI       #1{#1}\fi
\ifx \showISBNx    \undefined \def \showISBNx     #1{\unskip}     \fi
\ifx \showISBNxiii \undefined \def \showISBNxiii  #1{\unskip}     \fi
\ifx \showISSN     \undefined \def \showISSN      #1{\unskip}     \fi
\ifx \showLCCN     \undefined \def \showLCCN      #1{\unskip}     \fi
\ifx \shownote     \undefined \def \shownote      #1{#1}          \fi
\ifx \showarticletitle \undefined \def \showarticletitle #1{#1}   \fi
\ifx \showURL      \undefined \def \showURL       {\relax}        \fi
\providecommand\bibfield[2]{#2}
\providecommand\bibinfo[2]{#2}
\providecommand\natexlab[1]{#1}
\providecommand\showeprint[2][]{arXiv:#2}

\bibitem[\protect\citeauthoryear{Arefyev, Ermolaev, and Panchenko}{Arefyev
  et~al\mbox{.}}{2018}]%
        {arefyev2018much}
\bibfield{author}{\bibinfo{person}{Nikolay Arefyev}, \bibinfo{person}{Pavel
  Ermolaev}, {and} \bibinfo{person}{Alexander Panchenko}.}
  \bibinfo{year}{2018}\natexlab{}.
\newblock \showarticletitle{How much does a word weigh? Weighting word
  embeddings for word sense induction}.
\newblock \bibinfo{journal}{\emph{arXiv preprint arXiv:1805.09209}}
  (\bibinfo{year}{2018}).
\newblock


\bibitem[\protect\citeauthoryear{Bruna, Zaremba, Szlam, and LeCun}{Bruna
  et~al\mbox{.}}{2013}]%
        {bruna2013spectral}
\bibfield{author}{\bibinfo{person}{Joan Bruna}, \bibinfo{person}{Wojciech
  Zaremba}, \bibinfo{person}{Arthur Szlam}, {and} \bibinfo{person}{Yann
  LeCun}.} \bibinfo{year}{2013}\natexlab{}.
\newblock \showarticletitle{Spectral networks and locally connected networks on
  graphs}.
\newblock \bibinfo{journal}{\emph{arXiv preprint arXiv:1312.6203}}
  (\bibinfo{year}{2013}).
\newblock


\bibitem[\protect\citeauthoryear{Defferrard, Bresson, and
  Vandergheynst}{Defferrard et~al\mbox{.}}{2016}]%
        {defferrard2016convolutional}
\bibfield{author}{\bibinfo{person}{Micha{\"e}l Defferrard},
  \bibinfo{person}{Xavier Bresson}, {and} \bibinfo{person}{Pierre
  Vandergheynst}.} \bibinfo{year}{2016}\natexlab{}.
\newblock \showarticletitle{Convolutional neural networks on graphs with fast
  localized spectral filtering}.
\newblock \bibinfo{journal}{\emph{Advances in neural information processing
  systems}}  \bibinfo{volume}{29} (\bibinfo{year}{2016}),
  \bibinfo{pages}{3844--3852}.
\newblock


\bibitem[\protect\citeauthoryear{Di, Yu, Bu, and Sun}{Di et~al\mbox{.}}{2020}]%
        {di2020mutual}
\bibfield{author}{\bibinfo{person}{Xinhan Di}, \bibinfo{person}{Pengqian Yu},
  \bibinfo{person}{Rui Bu}, {and} \bibinfo{person}{Mingchao Sun}.}
  \bibinfo{year}{2020}\natexlab{}.
\newblock \showarticletitle{Mutual Information Maximization in Graph Neural
  Networks}. In \bibinfo{booktitle}{\emph{2020 International Joint Conference
  on Neural Networks (IJCNN)}}. IEEE, \bibinfo{pages}{1--7}.
\newblock


\bibitem[\protect\citeauthoryear{Eghbali and Tahvildari}{Eghbali and
  Tahvildari}{2019}]%
        {eghbali2019deep}
\bibfield{author}{\bibinfo{person}{Sepehr Eghbali} {and} \bibinfo{person}{Ladan
  Tahvildari}.} \bibinfo{year}{2019}\natexlab{}.
\newblock \showarticletitle{Deep spherical quantization for image search}. In
  \bibinfo{booktitle}{\emph{Proceedings of the IEEE/CVF Conference on Computer
  Vision and Pattern Recognition}}. \bibinfo{pages}{11690--11699}.
\newblock


\bibitem[\protect\citeauthoryear{Kim and Oh}{Kim and Oh}{2020}]%
        {kim2020find}
\bibfield{author}{\bibinfo{person}{Dongkwan Kim} {and} \bibinfo{person}{Alice
  Oh}.} \bibinfo{year}{2020}\natexlab{}.
\newblock \showarticletitle{How to find your friendly neighborhood: Graph
  attention design with self-supervision}. In
  \bibinfo{booktitle}{\emph{International Conference on Learning
  Representations}}.
\newblock


\bibitem[\protect\citeauthoryear{Kingma and Ba}{Kingma and Ba}{2014}]%
        {kingma2014adam}
\bibfield{author}{\bibinfo{person}{Diederik~P Kingma} {and}
  \bibinfo{person}{Jimmy Ba}.} \bibinfo{year}{2014}\natexlab{}.
\newblock \showarticletitle{Adam: A method for stochastic optimization}.
\newblock \bibinfo{journal}{\emph{arXiv preprint arXiv:1412.6980}}
  (\bibinfo{year}{2014}).
\newblock


\bibitem[\protect\citeauthoryear{Kipf and Welling}{Kipf and Welling}{2016a}]%
        {kipf2016semi}
\bibfield{author}{\bibinfo{person}{Thomas~N Kipf} {and} \bibinfo{person}{Max
  Welling}.} \bibinfo{year}{2016}\natexlab{a}.
\newblock \showarticletitle{Semi-supervised classification with graph
  convolutional networks}.
\newblock \bibinfo{journal}{\emph{arXiv preprint arXiv:1609.02907}}
  (\bibinfo{year}{2016}).
\newblock


\bibitem[\protect\citeauthoryear{Kipf and Welling}{Kipf and Welling}{2016b}]%
        {kipf2016variational}
\bibfield{author}{\bibinfo{person}{Thomas~N Kipf} {and} \bibinfo{person}{Max
  Welling}.} \bibinfo{year}{2016}\natexlab{b}.
\newblock \showarticletitle{Variational graph auto-encoders}.
\newblock \bibinfo{journal}{\emph{arXiv preprint arXiv:1611.07308}}
  (\bibinfo{year}{2016}).
\newblock


\bibitem[\protect\citeauthoryear{Klicpera, Bojchevski, and
  G{\"u}nnemann}{Klicpera et~al\mbox{.}}{2018}]%
        {klicpera2018predict}
\bibfield{author}{\bibinfo{person}{Johannes Klicpera},
  \bibinfo{person}{Aleksandar Bojchevski}, {and} \bibinfo{person}{Stephan
  G{\"u}nnemann}.} \bibinfo{year}{2018}\natexlab{}.
\newblock \showarticletitle{Predict then propagate: Graph neural networks meet
  personalized pagerank}.
\newblock \bibinfo{journal}{\emph{arXiv preprint arXiv:1810.05997}}
  (\bibinfo{year}{2018}).
\newblock


\bibitem[\protect\citeauthoryear{Kobayashi, Kuribayashi, Yokoi, and
  Inui}{Kobayashi et~al\mbox{.}}{2020}]%
        {kobayashi2020attention}
\bibfield{author}{\bibinfo{person}{Goro Kobayashi}, \bibinfo{person}{Tatsuki
  Kuribayashi}, \bibinfo{person}{Sho Yokoi}, {and} \bibinfo{person}{Kentaro
  Inui}.} \bibinfo{year}{2020}\natexlab{}.
\newblock \showarticletitle{Attention is not only a weight: Analyzing
  transformers with vector norms}.
\newblock \bibinfo{journal}{\emph{arXiv preprint arXiv:2004.10102}}
  (\bibinfo{year}{2020}).
\newblock


\bibitem[\protect\citeauthoryear{Li, Xiong, Thabet, and Ghanem}{Li
  et~al\mbox{.}}{2020}]%
        {li2020deepergcn}
\bibfield{author}{\bibinfo{person}{Guohao Li}, \bibinfo{person}{Chenxin Xiong},
  \bibinfo{person}{Ali Thabet}, {and} \bibinfo{person}{Bernard Ghanem}.}
  \bibinfo{year}{2020}\natexlab{}.
\newblock \showarticletitle{Deepergcn: All you need to train deeper gcns}.
\newblock \bibinfo{journal}{\emph{arXiv preprint arXiv:2006.07739}}
  (\bibinfo{year}{2020}).
\newblock


\bibitem[\protect\citeauthoryear{Liu, Wen, Yu, Li, Raj, and Song}{Liu
  et~al\mbox{.}}{2017}]%
        {liu2017sphereface}
\bibfield{author}{\bibinfo{person}{Weiyang Liu}, \bibinfo{person}{Yandong Wen},
  \bibinfo{person}{Zhiding Yu}, \bibinfo{person}{Ming Li},
  \bibinfo{person}{Bhiksha Raj}, {and} \bibinfo{person}{Le Song}.}
  \bibinfo{year}{2017}\natexlab{}.
\newblock \showarticletitle{Sphereface: Deep hypersphere embedding for face
  recognition}. In \bibinfo{booktitle}{\emph{Proceedings of the IEEE conference
  on computer vision and pattern recognition}}. \bibinfo{pages}{212--220}.
\newblock


\bibitem[\protect\citeauthoryear{Mavromatis and Karypis}{Mavromatis and
  Karypis}{2020}]%
        {mavromatis2020graph}
\bibfield{author}{\bibinfo{person}{Costas Mavromatis} {and}
  \bibinfo{person}{George Karypis}.} \bibinfo{year}{2020}\natexlab{}.
\newblock \showarticletitle{Graph InfoClust: Leveraging cluster-level node
  information for unsupervised graph representation learning}.
\newblock \bibinfo{journal}{\emph{arXiv preprint arXiv:2009.06946}}
  (\bibinfo{year}{2020}).
\newblock


\bibitem[\protect\citeauthoryear{Merrill, Ramanujan, Goldberg, Schwartz, and
  Smith}{Merrill et~al\mbox{.}}{2020}]%
        {merrill2020parameter}
\bibfield{author}{\bibinfo{person}{William Merrill}, \bibinfo{person}{Vivek
  Ramanujan}, \bibinfo{person}{Yoav Goldberg}, \bibinfo{person}{Roy Schwartz},
  {and} \bibinfo{person}{Noah Smith}.} \bibinfo{year}{2020}\natexlab{}.
\newblock \showarticletitle{Parameter Norm Growth During Training of
  Transformers}.
\newblock \bibinfo{journal}{\emph{arXiv preprint arXiv:2010.09697}}
  (\bibinfo{year}{2020}).
\newblock


\bibitem[\protect\citeauthoryear{Nguyen and Chiang}{Nguyen and Chiang}{2017}]%
        {nguyen2017improving}
\bibfield{author}{\bibinfo{person}{Toan~Q Nguyen} {and} \bibinfo{person}{David
  Chiang}.} \bibinfo{year}{2017}\natexlab{}.
\newblock \showarticletitle{Improving lexical choice in neural machine
  translation}.
\newblock \bibinfo{journal}{\emph{arXiv preprint arXiv:1710.01329}}
  (\bibinfo{year}{2017}).
\newblock


\bibitem[\protect\citeauthoryear{Nguyen and Salazar}{Nguyen and
  Salazar}{2019}]%
        {nguyen2019transformers}
\bibfield{author}{\bibinfo{person}{Toan~Q Nguyen} {and} \bibinfo{person}{Julian
  Salazar}.} \bibinfo{year}{2019}\natexlab{}.
\newblock \showarticletitle{Transformers without tears: Improving the
  normalization of self-attention}.
\newblock \bibinfo{journal}{\emph{arXiv preprint arXiv:1910.05895}}
  (\bibinfo{year}{2019}).
\newblock


\bibitem[\protect\citeauthoryear{Pan, Hu, Long, Jiang, Yao, and Zhang}{Pan
  et~al\mbox{.}}{2018}]%
        {pan2018adversarially}
\bibfield{author}{\bibinfo{person}{Shirui Pan}, \bibinfo{person}{Ruiqi Hu},
  \bibinfo{person}{Guodong Long}, \bibinfo{person}{Jing Jiang},
  \bibinfo{person}{Lina Yao}, {and} \bibinfo{person}{Chengqi Zhang}.}
  \bibinfo{year}{2018}\natexlab{}.
\newblock \showarticletitle{Adversarially regularized graph autoencoder for
  graph embedding}.
\newblock \bibinfo{journal}{\emph{arXiv preprint arXiv:1802.04407}}
  (\bibinfo{year}{2018}).
\newblock


\bibitem[\protect\citeauthoryear{Paszke, Gross, Massa, Lerer, Bradbury, Chanan,
  Killeen, Lin, Gimelshein, Antiga, et~al\mbox{.}}{Paszke
  et~al\mbox{.}}{2019}]%
        {paszke2019pytorch}
\bibfield{author}{\bibinfo{person}{Adam Paszke}, \bibinfo{person}{Sam Gross},
  \bibinfo{person}{Francisco Massa}, \bibinfo{person}{Adam Lerer},
  \bibinfo{person}{James Bradbury}, \bibinfo{person}{Gregory Chanan},
  \bibinfo{person}{Trevor Killeen}, \bibinfo{person}{Zeming Lin},
  \bibinfo{person}{Natalia Gimelshein}, \bibinfo{person}{Luca Antiga},
  {et~al\mbox{.}}} \bibinfo{year}{2019}\natexlab{}.
\newblock \showarticletitle{Pytorch: An imperative style, high-performance deep
  learning library}.
\newblock \bibinfo{journal}{\emph{Advances in neural information processing
  systems}}  \bibinfo{volume}{32} (\bibinfo{year}{2019}),
  \bibinfo{pages}{8026--8037}.
\newblock


\bibitem[\protect\citeauthoryear{Ranjan, Castillo, and Chellappa}{Ranjan
  et~al\mbox{.}}{2017}]%
        {ranjan2017l2}
\bibfield{author}{\bibinfo{person}{Rajeev Ranjan}, \bibinfo{person}{Carlos~D
  Castillo}, {and} \bibinfo{person}{Rama Chellappa}.}
  \bibinfo{year}{2017}\natexlab{}.
\newblock \showarticletitle{L2-constrained softmax loss for discriminative face
  verification}.
\newblock \bibinfo{journal}{\emph{arXiv preprint arXiv:1703.09507}}
  (\bibinfo{year}{2017}).
\newblock


\bibitem[\protect\citeauthoryear{Salha, Hennequin, and Vazirgiannis}{Salha
  et~al\mbox{.}}{2020}]%
        {salha2020simple}
\bibfield{author}{\bibinfo{person}{Guillaume Salha}, \bibinfo{person}{Romain
  Hennequin}, {and} \bibinfo{person}{Michalis Vazirgiannis}.}
  \bibinfo{year}{2020}\natexlab{}.
\newblock \showarticletitle{Simple and effective graph autoencoders with
  one-hop linear models}.
\newblock \bibinfo{journal}{\emph{arXiv preprint arXiv:2001.07614}}
  (\bibinfo{year}{2020}).
\newblock


\bibitem[\protect\citeauthoryear{Schakel and Wilson}{Schakel and
  Wilson}{2015}]%
        {schakel2015measuring}
\bibfield{author}{\bibinfo{person}{Adriaan~MJ Schakel} {and}
  \bibinfo{person}{Benjamin~J Wilson}.} \bibinfo{year}{2015}\natexlab{}.
\newblock \showarticletitle{Measuring word significance using distributed
  representations of words}.
\newblock \bibinfo{journal}{\emph{arXiv preprint arXiv:1508.02297}}
  (\bibinfo{year}{2015}).
\newblock


\bibitem[\protect\citeauthoryear{Shchur, Mumme, Bojchevski, and
  G{\"u}nnemann}{Shchur et~al\mbox{.}}{2018}]%
        {shchur2018pitfalls}
\bibfield{author}{\bibinfo{person}{Oleksandr Shchur},
  \bibinfo{person}{Maximilian Mumme}, \bibinfo{person}{Aleksandar Bojchevski},
  {and} \bibinfo{person}{Stephan G{\"u}nnemann}.}
  \bibinfo{year}{2018}\natexlab{}.
\newblock \showarticletitle{Pitfalls of graph neural network evaluation}.
\newblock \bibinfo{journal}{\emph{arXiv preprint arXiv:1811.05868}}
  (\bibinfo{year}{2018}).
\newblock


\bibitem[\protect\citeauthoryear{Veli{\v{c}}kovi{\'c}, Cucurull, Casanova,
  Romero, Lio, and Bengio}{Veli{\v{c}}kovi{\'c} et~al\mbox{.}}{2017}]%
        {velivckovic2017graph}
\bibfield{author}{\bibinfo{person}{Petar Veli{\v{c}}kovi{\'c}},
  \bibinfo{person}{Guillem Cucurull}, \bibinfo{person}{Arantxa Casanova},
  \bibinfo{person}{Adriana Romero}, \bibinfo{person}{Pietro Lio}, {and}
  \bibinfo{person}{Yoshua Bengio}.} \bibinfo{year}{2017}\natexlab{}.
\newblock \showarticletitle{Graph attention networks}.
\newblock \bibinfo{journal}{\emph{arXiv preprint arXiv:1710.10903}}
  (\bibinfo{year}{2017}).
\newblock


\bibitem[\protect\citeauthoryear{Wang, Xiang, Cheng, and Yuille}{Wang
  et~al\mbox{.}}{2017}]%
        {wang2017normface}
\bibfield{author}{\bibinfo{person}{Feng Wang}, \bibinfo{person}{Xiang Xiang},
  \bibinfo{person}{Jian Cheng}, {and} \bibinfo{person}{Alan~Loddon Yuille}.}
  \bibinfo{year}{2017}\natexlab{}.
\newblock \showarticletitle{Normface: L2 hypersphere embedding for face
  verification}. In \bibinfo{booktitle}{\emph{Proceedings of the 25th ACM
  international conference on Multimedia}}. \bibinfo{pages}{1041--1049}.
\newblock


\bibitem[\protect\citeauthoryear{Wang, Wang, Zhou, Ji, Gong, Zhou, Li, and
  Liu}{Wang et~al\mbox{.}}{2018}]%
        {wang2018cosface}
\bibfield{author}{\bibinfo{person}{Hao Wang}, \bibinfo{person}{Yitong Wang},
  \bibinfo{person}{Zheng Zhou}, \bibinfo{person}{Xing Ji},
  \bibinfo{person}{Dihong Gong}, \bibinfo{person}{Jingchao Zhou},
  \bibinfo{person}{Zhifeng Li}, {and} \bibinfo{person}{Wei Liu}.}
  \bibinfo{year}{2018}\natexlab{}.
\newblock \showarticletitle{Cosface: Large margin cosine loss for deep face
  recognition}. In \bibinfo{booktitle}{\emph{Proceedings of the IEEE conference
  on computer vision and pattern recognition}}. \bibinfo{pages}{5265--5274}.
\newblock


\bibitem[\protect\citeauthoryear{Wu, Souza, Zhang, Fifty, Yu, and
  Weinberger}{Wu et~al\mbox{.}}{2019}]%
        {wu2019simplifying}
\bibfield{author}{\bibinfo{person}{Felix Wu}, \bibinfo{person}{Amauri Souza},
  \bibinfo{person}{Tianyi Zhang}, \bibinfo{person}{Christopher Fifty},
  \bibinfo{person}{Tao Yu}, {and} \bibinfo{person}{Kilian Weinberger}.}
  \bibinfo{year}{2019}\natexlab{}.
\newblock \showarticletitle{Simplifying graph convolutional networks}. In
  \bibinfo{booktitle}{\emph{International conference on machine learning}}.
  PMLR, \bibinfo{pages}{6861--6871}.
\newblock


\bibitem[\protect\citeauthoryear{Wu, Guo, Suresh, Kumar, Holtmann-Rice, Simcha,
  and Yu}{Wu et~al\mbox{.}}{2017}]%
        {wu2017multiscale}
\bibfield{author}{\bibinfo{person}{Xiang Wu}, \bibinfo{person}{Ruiqi Guo},
  \bibinfo{person}{Ananda~Theertha Suresh}, \bibinfo{person}{Sanjiv Kumar},
  \bibinfo{person}{Daniel~N Holtmann-Rice}, \bibinfo{person}{David Simcha},
  {and} \bibinfo{person}{Felix Yu}.} \bibinfo{year}{2017}\natexlab{}.
\newblock \showarticletitle{Multiscale quantization for fast similarity
  search}.
\newblock \bibinfo{journal}{\emph{Advances in Neural Information Processing
  Systems}}  \bibinfo{volume}{30} (\bibinfo{year}{2017}),
  \bibinfo{pages}{5745--5755}.
\newblock


\bibitem[\protect\citeauthoryear{Xu, Shen, Cao, Cen, and Cheng}{Xu
  et~al\mbox{.}}{2020}]%
        {xu2020graph}
\bibfield{author}{\bibinfo{person}{Bingbing Xu}, \bibinfo{person}{Huawei Shen},
  \bibinfo{person}{Qi Cao}, \bibinfo{person}{Keting Cen}, {and}
  \bibinfo{person}{Xueqi Cheng}.} \bibinfo{year}{2020}\natexlab{}.
\newblock \showarticletitle{Graph convolutional networks using heat kernel for
  semi-supervised learning}.
\newblock \bibinfo{journal}{\emph{arXiv preprint arXiv:2007.16002}}
  (\bibinfo{year}{2020}).
\newblock


\bibitem[\protect\citeauthoryear{Zhang, Li, and Zhang}{Zhang
  et~al\mbox{.}}{2020}]%
        {zhang2020deep}
\bibfield{author}{\bibinfo{person}{Dingyi Zhang}, \bibinfo{person}{Yingming
  Li}, {and} \bibinfo{person}{Zhongfei Zhang}.}
  \bibinfo{year}{2020}\natexlab{}.
\newblock \showarticletitle{Deep Metric Learning with Spherical Embedding}.
\newblock \bibinfo{journal}{\emph{Advances in Neural Information Processing
  Systems}}  \bibinfo{volume}{33} (\bibinfo{year}{2020}).
\newblock


\bibitem[\protect\citeauthoryear{Zhao and Akoglu}{Zhao and Akoglu}{2019}]%
        {zhao2019pairnorm}
\bibfield{author}{\bibinfo{person}{Lingxiao Zhao} {and} \bibinfo{person}{Leman
  Akoglu}.} \bibinfo{year}{2019}\natexlab{}.
\newblock \showarticletitle{Pairnorm: Tackling oversmoothing in gnns}.
\newblock \bibinfo{journal}{\emph{arXiv preprint arXiv:1909.12223}}
  (\bibinfo{year}{2019}).
\newblock


\bibitem[\protect\citeauthoryear{Zheng, Pal, and Savvides}{Zheng
  et~al\mbox{.}}{2018}]%
        {zheng2018ring}
\bibfield{author}{\bibinfo{person}{Yutong Zheng}, \bibinfo{person}{Dipan~K
  Pal}, {and} \bibinfo{person}{Marios Savvides}.}
  \bibinfo{year}{2018}\natexlab{}.
\newblock \showarticletitle{Ring loss: Convex feature normalization for face
  recognition}. In \bibinfo{booktitle}{\emph{Proceedings of the IEEE conference
  on computer vision and pattern recognition}}. \bibinfo{pages}{5089--5097}.
\newblock


\end{thebibliography}

\appendix

\end{document}